\documentclass{ICRES}
\usepackage{graphicx}
\usepackage{xcolor}
\usepackage{url}
\usepackage[utf8]{inputenc}
\makeatletter
\g@addto@macro{\UrlBreaks}{\UrlOrds}
\makeatother

\begin{document}

\title{A Simulated real-world upper-body Exoskeleton Accident and Investigation}
%
%\titlerunning{EBB Open Standard}  % abbreviated title (for running head)
%                                     also used for the TOC unless
%                                     \toctitle is used
%
\author{Alan F.T. Winfield, Nicola Webb and Appolinaire Etoundi}
\address{Bristol Robotics Lab, University of the West of England\\ 
Bristol, BS16 1QY, UK\\
E-mail: alan.winfield@brl.ac.uk}

\author{Romain Derval}
\address{Tribonix Ltd, Room 2x102, Bristol Business School, Frenchay Campus, Coldharbour Lane\\
Bristol, BS16 1QY UK}

\author{Pericle Salvini and Marina Jirotka}
\address{Department of Computer Science, University of Oxford\\ Oxford, UK}
%
%\authorrunning{Alan Winfield} % abbreviated author list (for running head)
%
%%%% list of authors for the TOC (use if author list has to be modified)
%\tocauthor{Alan Winfield}
%

%\maketitle              % typeset the title of the contribution

\begin{abstract}
This paper describes the enactment of a simulated (mock) accident involving an upper-body exoskeleton and its investigation. The accident scenario is enacted by role-playing volunteers, one of whom is wearing the exoskeleton. Following the mock accident, investigators -- also volunteers -- interview both the subject of the accident and relevant witnesses. The investigators then consider the witness testimony alongside robot data logged by the ethical black box, in order to address the three key questions: \textit{what} happened?, \textit{why} did it happen?, and \textit{how} can we make changes to prevent the accident happening again? This simulated accident scenario is one of a series we have run as part of the RoboTIPS project, with the overall aim of developing and testing both processes and technologies to support social robot accident investigation.
\end{abstract}

\bodymatter

\section{Introduction}
Accidents are inevitable and this is just as true for robots as any other machinery. Accidents involving social robots -- robots operating in close proximity to humans -- are much more likely than for industrial robots, as the latter are usually installed within safety cages. In \cite{winf2021} we argued that both processes and technologies to support social robot accident investigation are needed. We have proposed and developed a key technology needed to underpin accident or incident (near miss) investigations: a data logger which we refer to as an Ethical Black Box (EBB) \cite{winf2017,Winf2022}. Also related to the present work we have investigated the engagement of role play and real human witnesses within a virtual accident scenario \cite{Webb2021}. 

The simulation of accidents in the real-world is commonplace. Consider for instance vehicle crash tests including famously a train collision with a nuclear flask \cite{Miles1987}. Of course none of these simulations involve humans, except as observers, although the use of crash test dummies is well established \cite{Fost1977}. Karwowski et at \cite{Karw1991} describe the real-world simulation of industrial robot accidents, making use of a manikin as a proxy human; the aim of this study was to investigate the impact of simulated accidents on real human worker safety training. In Human Robot Interaction (HRI) computer simulation has been used to study HRI accidents involving virtual robots and humans \cite{mans2013}.

These accident simulations -- both real-world and virtual -- have typically not simulated the process of accident investigation. Accident investigation is a \textit{human} process that relies on both technical data (ideally from a data logger) and testimony from both witnesses and experts. But witness accounts of what happened cannot be collected from either real-world manikins or virtual humans. It is for this reason that, in our work, we eschew computer simulation and instead simulate the accident and its investigation in the real-world. We believe this to be the first time this approach has been applied in human robot interaction.

This paper is structured as follows. Section 2  outlines the experimental method we have developed for staging mock robot accidents and investigations. In Section 3 we introduce the field of wearable robots and exoskeletons, before then describing the Tribonix\footnote{https://tribonix.com/} upper-body exoskeleton utilised in this work. Section 4 details the accident scenario developed for the upper-body exoskeleton, then -- in Section 5 -- we describe both the enactment of the accident scenario, and its subsequent investigation. Section 6 concludes the paper.

\section{Experimental method}

The approach to the real-world simulation of social robot accidents and investigation that we have developed is human centric. The key elements of our experimental method are:

\begin{enumerate}
    \item The accident scenario is enacted by human volunteers, role playing the subject of the accident, together with both direct  and indirect  witnesses. The subject is the person to whom the accident happens. Direct witnesses are those who either witness or discover the accident, and indirect witnesses are those who might be supervisors or managers of the subject and/or the facility, or representatives of the robot's manufacturer.
    \item Prior to the enactment the project team brief the volunteers. Each briefing is specific to the role and, with the exception of the subject, volunteers are briefed only on their role, and not the whole scenario. This is so that they witness the accident (or it's aftermath) for the first time during the enactment. Only the subject is fully briefed on the scenario, including the safety aspects explained below, so that they are confident that they will not come to harm or be fearful during the enactment.
    \item The enactment is stage managed by project team members. Although the simulation resembles a piece of theatre, volunteers are not asked to learn any lines. Apart from any specific action essential to the scenario (which will be prompted by the stage manager) the volunteers are invited to \textit{ad lib} in a way that is appropriate to the roles they are playing. Volunteers are asked to wait in a side room until they are called a few moments before they are needed.
    \item Safety of the volunteers, and especially the subject, is of paramount importance. Thus, if the scenario simulates physical harm to the subject, then -- when the accident happens -- the enactment is briefly suspended by the stage manager and the subject is helped into the position they might be expected to be in, following the accident. The project team conduct a safety risk assessment and if necessary modify the scenario and/or its stage management to mitigate any risks and the simulation is only undertaken after university research ethics approval.
    \item The accident investigators are also volunteers and, ideally, the lead accident investigator has expertise and/or experience in accident investigation. Robotics expertise is not essential, as the aims and process of investigation are common to all accident or incident (near miss) investigations. The accident investigators are not briefed on the scenario, only the type of robot involved. Necessarily the accident investigators are not present during the enactment of the simulated accident. To reduce the time burden on all volunteers we stage the accident and its investigation on a single day, with the accident investigators arriving after the enactment.
\end{enumerate}

It is important to note that although the accident is a simulation enacted under controlled conditions, its investigation is \textit{not} a simulation. For all intents and purposes our investigators approach their task as if they are investigating a real accident. Of course there are several aspects of the investigation that are unrealistic, for instance (i) a real investigation would take place over weeks or months and (ii) a real investigation team would be larger than our two people.
%and (iii) the findings of the investigation of a mock accident has limited direct real-world consequence. 
But, aside from these aspects, our investigation follows the same process, uses the same kinds of evidence -- witness accounts and data logs -- and addresses the same three questions of all accident investigations, namely: what happened?, why did it happen? and what can we do to prevent it happening again?

\section{The Exoskeleton}

\subsection{Introduction to Wearable Robots and Exoskeletons}

\paragraph{Evolution and Definition}
The trajectory of robotics development has increasingly intertwined with human interaction. Historically, robots were deployed primarily within industrial settings, tasked with automating repetitive or precision-required activities traditionally performed by humans. However, the evolution of robotics has shifted towards a paradigm where human-robot interaction is not merely functional but integrative, spanning both physical and cognitive realms \cite{Alami2006}.

This shift marks the genesis and growing importance of wearable robots, which characterize the blending of human capabilities with robotic technology. Wearable robots, tailored to be ‘worn’ by users, either augment the functionalities of human limbs or serve as replacements for them. These devices range in application from orthotic robots and exoskeletons, which assist or enhance limb function, to prosthetic devices that substitute for limbs lost to amputation.

The definition of wearability in robots transcends the notions of mobility, portability, or autonomy. In some cases where wearable robots remain non-ambulatory, this limitation is often due to current constraints in technology, particularly in the areas of actuators and power sources. Fundamentally, wearable robots are designed to extend, complement, substitute, or amplify human physical capabilities, offering new levels of empowerment or functional replacement to the areas of the body they are equipped on. This development not only reflects the technical progression in robotics but also highlights a significant shift towards more personal and physically integrated robotic applications.

Wearable robots, particularly exoskeletons, represent a groundbreaking frontier in modern robotics, merging human biomechanics with robotic technology to enhance, augment, or restore human performance. The concept of exoskeletons is not merely futuristic; it can be traced back to early patents and literary mentions within science fiction realms, but substantive research kicked off in the late 1960s \cite{Hein1959}. During this period, development efforts emerged almost simultaneously but were distinctly focused across different geopolitical landscapes — in the United States and former Yugoslavia. In the U.S., the primary objective revolved around augmenting the abilities of able-bodied individuals, predominantly for military applications. Contrastingly, Yugoslavian efforts were more aligned with creating assistive technologies designed to aid physically challenged individuals. Modern exoskeletons now span applications from medical rehabilitation to industrial support and even personal mobility assistance.

An exoskeleton is defined as an anthropomorphic, active mechanical device that is closely ``worn" by an operator, complementing and amplifying the operator’s movements. These devices can be powered by a variety of means, including electric motors, pneumatics, levers, and hydraulics, and are controlled through a combination of computer processing, human input, and sometimes autonomous sensors \cite{Pons2008}.

\paragraph{Technological Integration and Human-Machine Interaction}
The integration of sensors and advanced control systems in exoskeletons facilitates a sophisticated human-machine interface (HMI). This interaction is crucial not only for the intuitive operation of the device but also for ensuring the safety and comfort of the user. The control systems in these robots often employ algorithms based on machine learning and real-time feedback to adapt to individual users’ needs \cite{gopu2016}.

\subsubsection{Applications Across Fields}

\paragraph{Medical Rehabilitation}
In medical settings, exoskeletons are revolutionary, providing mobility solutions for individuals suffering from paralysis, muscular diseases, or skeletal injuries. These devices assist in not only enabling movement but also in therapeutic roles, aiding in the recovery and strengthening of bodily functions \cite{Esqu2012}. They offer new horizons in physical therapy, where repetitive and assistive movements facilitated by robotic systems can lead to faster recovery and rehabilitation.

\paragraph{Industrial and Military Enhancement}
Industrial exoskeletons are designed to support workers performing physically demanding tasks, reducing fatigue and preventing workplace injuries. This application underscores a significant ethical consideration: the enhancement of human capabilities in work environments must balance productivity gains with workers' rights and safety standards \cite{Looz2016}.

Similarly, military exoskeletons aim to enhance soldier performance while reducing the risk of injuries. These applications often push the boundaries of what is technically feasible, driving innovations that eventually benefit civilian applications \cite{Boge2009}.

\subsubsection{Ethical, Legal, and Societal Considerations}

\paragraph{Ethical and Privacy Concerns}
The integration of exoskeletons into daily life brings forth complex ethical questions. Issues such as privacy, especially concerning data collected by these devices, informed consent, and autonomy must be meticulously managed. The potential for continuous monitoring raises significant privacy concerns, necessitating robust data management and protection frameworks \cite{Calo2011}.

\paragraph{Autonomy and Liability}
The question of autonomy and decision-making in conjunction with exoskeleton use is pivotal. As these systems incorporate more autonomous features, determining liability in the case of malfunctions or accidents becomes complex. This ties into the broader discussion of AI ethics and the role of intelligent systems in shared decision-making processes \cite{matt2004}.

\paragraph{Standardisation and Safety}
Standardisation of safety protocols for exoskeletons is still in development. Unlike traditional industrial machinery or medical devices, the intimate interaction of exoskeletons with the human body presents unique challenges in standardisation. Ensuring consistent safety across different systems and applications requires international collaboration and agreement on safety and performance metrics \cite{ISO299}.

\subsection{The Tribonix Upper-Body Exoskeleton}

\begin{figure}[ht]
\centering
\includegraphics[width=5.0cm]{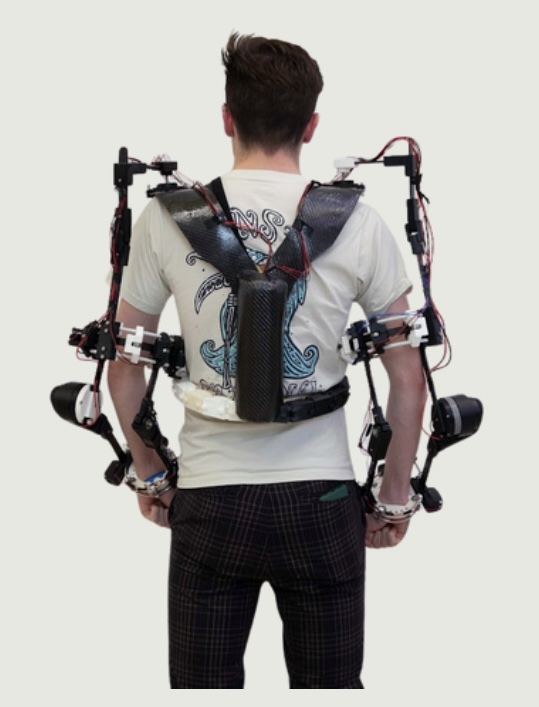}
\caption{The Tribonix Upper Body Exoskeleton (photo: Tribonix Ltd)}
\label{Exo}
\end{figure} 

The Tribonix exoskeleton, shown in Fig. \ref{Exo}, has a mechanism designed to allow forearm pronation/supination, internal medial rotation and external medial rotation of the arm. whilst empowering the users in standard arm movements. This system is voice-controlled for opening and closing, facilitating the process of donning and doffing for users, complemented by electromyography for seamless movement. These features are a major improvement over traditional exoskeleton design. The device attaches to the user's biceps and forearm, incorporating a circular mechanism that acts as rails for the opening/closing system. This mechanism can open, allowing the user to slide their arm in easily. Once secured, it enables free and independent rotation within the device's structure. The total weight of the exoskeleton is 6.2 kilos making it light to wear. It allows for 94\% of the human arm movement. Designed for rehabilitation, the flexion extension of the arm is powered to help the user to carry up to 6 kilos plus arm weight. In future all the movements of the arm will be powered.

In preparation for the accident simulation we developed and tested an ethical black box (EBB) for the Tribonix exoskeletion. Sensor and actuator data is captured by a microcontroller embedded in the exoskeleton and transmitted via Bluetooth to a laptop, where the data is formatted and stored in a CSV file. The EBB captures 12 datum points once per second. The EBB data consists of 2 groups. The first group of 6 data is EMG (electromyography) sensors and decisions. For each of the wearer's left and right tricep and bicep, there are EMG sensor readings plus a decision signal which activates the exoskeleton's left and right elbow actuators. The second group of 6 data captures the left and right elbow position, torque and temperature.

\section{The mock accident Scenario}

\paragraph{The Cast} 
\begin{itemize}
    \item Paul. Exoskeleton wearer and subject of the accident
    \item Mike. Co-worker and supervisor 1
    \item Supervisor 2
    \item Senior Manager
    \item The Paramedic
    \item The Expert and representative of the manufacturer
\end{itemize}

\paragraph{Synopsis} 
Paul and Mike (all names are fictional) are responsible for the storage of objects in a warehouse. Paul is wearing a powered suit, an exoskeleton designed to ease manual workers daily lifting and carrying tasks, reducing muscle strain and postponing fatigue. Paul’s task is to lift heavy containers and place them in high shelves, while Mike’s task is to both scan each object label with his tablet to check the correct storage position, and supervise Paul’s activities. One day, during working hours, they have an argument which soon turns into a scuffle in which Mike comes off worst: he falls and hurts his back. 

Alerted by the noise, colleagues arrive, separate the two men and calm them down. According to Mike, Paul has used the ``augmented” strength given by his exoskeleton to push him on purpose. On the contrary, Paul maintains that his colleague was goading him, trying to throw him off balance, but, anyway, he insists that he did not push Mike.

With a backache, Mike goes home to rest whereas Paul decides to resume his task, notwithstanding his manager and colleagues’ strong suggestion to take a break. Without waiting for a replacement supervisor to arrive, Paul starts lifting and shelving boxes on his own. He wants to finish his job without further delay. However, an accident happens. An box falls from his hands and hits his face. Moments later supervisor  2 arrives and finds Paul lying on the floor, senseless, and raises the alarm.

Paul's colleagues (senior manager and supervisor 2) perform first aid and wait for the ambulance. When the para-medic arrives, he cannot move Paul’s hurt arm because of the exoskeleton stiffness. Moreover, it takes some time to remove the exoskeleton from Paul to perform health checks. At the hospital, besides bruises on his face, doctors find many bruises on his upper body.

According to Paul’s version of events, he was lifting a box and almost reached the shelf when all of a sudden, the box fell on him. He tried to protect his face with his hands, but he couldn't because the exoskeleton prevented that movement. The device was not responding to his movements as usual. More than this he could not remember because he became unconscious. 

According to the senior manager, Paul was very nervous following the scuffle with Mike and they think his emotional state affected his abilities to control the exoskeleton. It is possible that he performed the wrong movement which caused him to lose grip with the object. He should have stopped working that day, as they suggested.

\section{Enactment and Investigation}

\subsection{The mock accident}

The scenario outlined above was enacted by volunteers in February 2024. The enactment was performed in the Bristol Robotics Laboratory in a room prepared with shelving and boxes. Project team members, including Tribonix staff, were present to not only stage manage the simulation but also carefully monitor all technical and safety aspects.

\begin{figure}[htbp]
\centering
\includegraphics[width=12.0cm]{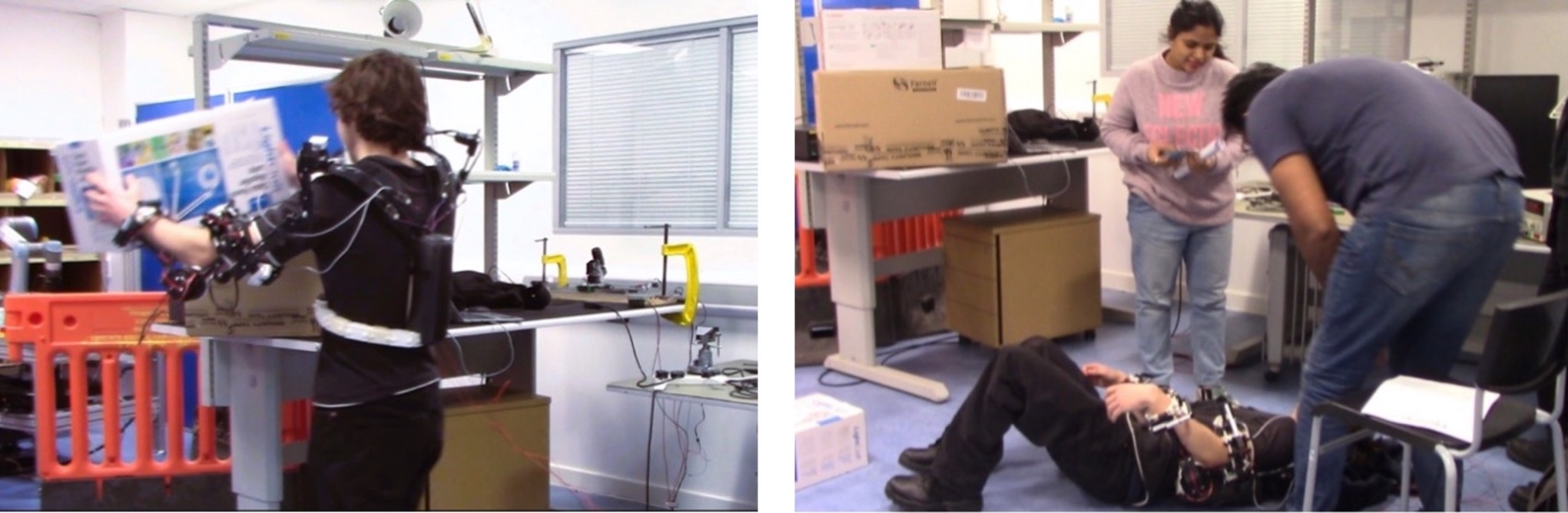}
\caption{Left: Paul lifting boxes. Right: Paul lying on the floor attended by the paramedic (right) and senior manager}
\label{Fig1}
\end{figure} 

The simulation was successfully enacted, with all volunteers splendidly playing their parts. Voice control of the exoskeleton was not used. During the simulation live EBB data was captured from the exoskeleton. Unfortunately a minor fault on the day of the simulation meant that data from the EMG sensors was not captured by the EBB. In all other respects the exoskeleton was operating normally. 

\subsection{The accident Investigation}

\begin{figure}[ht]
\centering
\includegraphics[width=7.0cm]{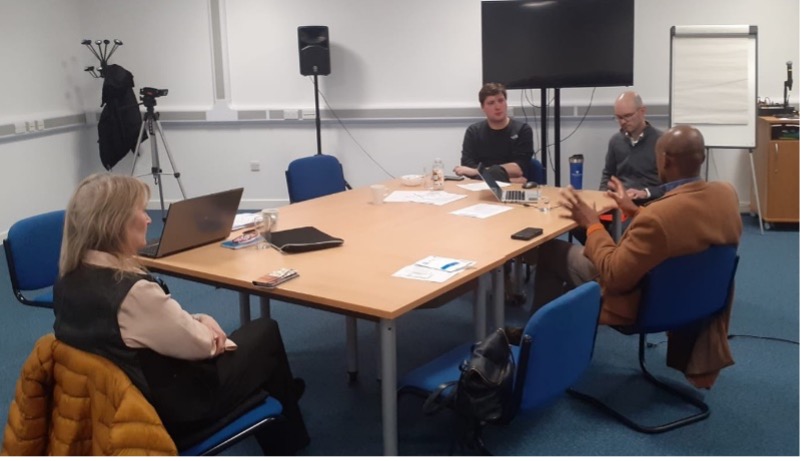}
\caption{The accident investigators interviewing the representative of the robot manufacturer (right), observed by a project team member (left).}
\label{Fig2}
\end{figure} 

Our two accident investigators first inspected the scene of the accident and the exoskeleton. They then interviewed the subject and all direct and indirect witnesses, including Paul, Mike, supervisor 2, the senior manager, the paramedic and the representative of the robot manufacturer. They then examined the data logs from the exoskeleton’s ethical black box (EBB), before pulling all of the evidence together to draw their conclusions.

From their discussions with the witnesses, the accident investigators quickly realised that they were investigating not one but two accidents: the first was Mike's fall following the argument between Paul and Mike, and the second -- more serious accident -- was Paul's fall while operating the exoskeleton unsupervised.

\paragraph{Accident 1: The scuffle} Interviews with Paul and Mike established the fact of the argument and the ensuing physical contact between the two men. The investigators were, however, unable to determine whether Paul pushed Mike, or Mike simply tripped during the scuffle. The investigators were able to deduce from examination of the EBB data that Paul might not have pushed pushed Mike, as the data did not show exoskeleton arm movements at the time of the scuffle. However, the lack of EMG sensor data meant that the investigators could not be confident of this.

\paragraph{Accident 2: Paul's fall} It was relatively easy for the investigators to determine the facts of the events leading up to Paul's fall while operating the exoskeleton. First was Paul's ill advised decision to ignore his manager's advice to take a break from work, compounded by his clear breach of company safety protocols by not waiting until a replacement supervisor arrived. A second -- and key finding of the investigation -- was that the exoskeleton suffered a complete loss of power while Paul was lifting a box. The EBB data showed that, at the time of Paul's fall, all data points went to zero, providing clear and unambiguous evidence of the power loss. Note that the disconnection of power to the exoskeleton was a planned part of the scenario, that was carefully done while Paul was not in the process of moving a box.\\ 

Thus our investigators were able to establish what happened and why, with low confidence for accident 1 and high confidence for accident 2. The investigators were also able to suggest a number of recommendations for preventing the same kinds of accidents happening again:
\begin{enumerate}
    \item Operational recommendation: minimum safe distance of co-workers must be maintained at all times around the exoskeleton wearer.
    \item Operational recommendation: instructions from a senior manager to suspend work and/or wait for a supervisor before resuming work should be mandatory.
    \item Technical recommendation: a fail safe system should be devised and built into the exoskeleton to ensure that power failure cannot cause the exo to freeze while under load.
    \item Technical recommendation: an alarm system should be built into the exoskeleton to alert the user if EMG sensors have failed.
\end{enumerate}

\section{Discussion and Conclusions}

There are two overall findings of this experimental study: (i) there is considerable value in real-world simulation of social robot accidents, and their investigation, both for the exploration of how human-robot interaction can lead to accidents, or near-miss incidents, and how operational and/or technical changes might reduce the likelihood of such accidents, and (ii) the critical role of the EBB in providing the data to support the accident investigation.

The study also led to the following observations:

\begin{itemize}
    \item \textit{Expect the unexpected}. Even with careful planning and controlled conditions real-world simulations give rise to unexpected factors. One such factor was the unavailability of data from the exoskeleton's EMG sensors. But we should treat this not as a problem, but simply a consequence of the real-world approach we have adopted.
    \item \textit{Repeatability} Another consequence of our real-world approach to accident simulation and investigation is that we do not have the strict repeatability expected of scientific experimentation. We would counter this with the observation that the same is true for any human-robot experiments, involving real robots and real humans.
    \item \textit{Investigator expertise} Our 2-person accident investigation team expressed difficulty in two areas. The first was that the team did not include an expert on exoskeletons. A real accident investigation team would normally include a domain expert, who would also be able to liaise with the manufacturer to ask specific questions of detail arising from the data log. The second and related difficulty was interpretation of the EBB data. Investigators found it hard to understand the data collected from the EMG sensors and control system. They remarked that an EBB visualisation tool would be a great help if, for instance, it offered different ways of accessing particular data items, (i.e. plotting trends over time), or the ability to play back (or animate) the EBB data.
    \item \textit{Order of investigation} In de-briefing our investigators expressed the view that, in retrospect, they should have examined the EBB data prior to interviewing witnesses. By so doing they might have asked different and more directed questions. However, investigators also pointed out that there is a risk in starting with examination of the EBB data. Data logs can be persuasive and potentially misleading since they tell only part of the story. %In general investigators recommended drawing on diverse sources of evidence (e.g., scene inspections, witness testimony, etc).
    \item \textit{Recommendation for an EBB heartbeat} It also emerged during the de-briefing discussion that the EBB would benefit from a `heartbeat' record, which simply means `robot powered up and working'. This would remove the ambiguity from EBB records which are all zero, that could be interpreted as either a robot failure or an EBB fault.
\end{itemize}

In common with all accidents, robot accidents are often outcomes of complex interacting factors, including both human-human and human-robot interactions, combined with both operational and technical failures. If humans are to have confidence in robots then transparent processes of accident investigation, and subsequent learning, will need to become part of the landscape of safe and ethical social robotics.

\section*{Acknowledgments}

We are extremely grateful to the student volunteers who took part in the simulation: Dan Read, Ashwin Aravind Chandapur, Monica Monica, Surin Machaiah and Ben Allen. The RoboTIPS project team are equally indebted to Dr Appolinaire Etoundi who played the role of expert representative of the manufacturer, Professor Carl Macrae who led the accident investigation and his co-investigator Jack Hughes. The project team are especially grateful to Romain Derval and Filip Hanus, co-founders of Tribonix, for both kindly agreeing to the use of their exoskeleton and generously working with us during the planning and enactment of this simulation. The simulation was undertaken with Research Ethics Approval CATE-2324-218, and the work was conducted within project RoboTIPS: \textit{Developing Responsible Robots for the Digital Economy} supported by EPSRC grant ref EP/S005099/1.

\bibliographystyle{ICRES}
\bibliography{EBBpaper}

\end{document}